\documentclass{article}
\usepackage{spconf,amsmath,graphicx,amsfonts}
\usepackage{lineno}
\usepackage{enumerate}
\usepackage{setspace,url,float}
\usepackage{lscape,multicol,multirow,longtable,tabularx}
\usepackage{algorithm}
\usepackage{algpseudocode}
\usepackage{array}
\newcolumntype{P}[1]{>{\centering\arraybackslash}p{#1}}
\title{PADA: Pruning Assisted Domain Adaptation for Self-Supervised Speech Representations}
\name{Vasista Sai Lodagala$^1$, Sreyan Ghosh$^2$, S. Umesh$^1$}
\address{$^1$Indian Institute of Technology, Madras\\  $^2$University of Maryland, College Park}
\copyrightnotice{978-1-6654-7189-3/22/\$31.00~\copyright2023 IEEE}
\begin{document}
%\ninept
%
\maketitle
\begin{abstract}
%While self-supervised speech representation learning (SSL) models serve a variety of downstream tasks, these models have been observed to overfit to the domain from which the unlabeled data originates. To alleviate this issue, we propose \textbf{PADA} (\textbf{P}runing \textbf{A}ssisted \textbf{D}omain \textbf{A}daptation), and zero out redundant weights from models pre-trained on large amounts of \emph{out-of-domain} (OOD) data. Intuitively, this helps to make space for the \emph{target-domain} ASR finetuning. The redundant weights can be identified through various pruning strategies which have been discussed in detail as a part of this work. Specifically, we investigate the effect of the recently discovered \emph{Task-Agnostic} and \emph{Task-Aware} pruning on PADA and propose a new pruning paradigm based on the latter, which we call \emph{Cross-Domain Task-Aware Pruning} (CD-TAW). CD-TAW obtains the initial pruning mask from a well fine-tuned OOD model, which makes it starkly different from the rest of the pruning strategies discussed in the paper. Our proposed CD-TAW methodology achieves up to 20.6\% relative WER improvement over our baseline when fine-tuned on a 2-hour subset of Switchboard data without language model (LM) decoding. Furthermore, we conduct detailed analysis to highlight the key design choices of our proposed method.
While self-supervised speech representation learning (SSL) models serve a variety of downstream tasks, these models have been observed to overfit to the domain from which the unlabeled data originates. To alleviate this issue, we propose \textbf{PADA} (\textbf{P}runing \textbf{A}ssisted \textbf{D}omain \textbf{A}daptation). Before performing the \emph{target-domain} ASR fine-tuning, we discover the redundant weights from pre-trained wav2vec 2.0 models through various pruning strategies. We investigate the effect of \emph{Task-Agnostic} and \emph{Task-Aware} pruning and propose a new pruning paradigm called, \emph{Cross-Domain Task-Aware Pruning} (CD-TAW). CD-TAW obtains the initial pruning mask from a well fine-tuned \emph{out-of-domain} (OOD) model, thereby making use of the readily available fine-tuned models from the web. The proposed CD-TAW method achieves up to 20.6\% relative WER improvement over our baseline when fine-tuned on a 2-hour subset of Switchboard data without language model (LM) decoding. 
%Furthermore, we conduct a detailed analysis to highlight the key design choices of our proposed method. 
\end{abstract}

\section{Introduction}

Over the past decade, Automatic Speech Recognition (ASR) has drawn the attention of researchers from various fields owing to its potential applications in various Natural Language Understanding (NLU) systems having speech as the primary modality of communication \cite{gaikwad2010review, ney1999speech, saon2017english, yadav2020end}. The advent of Deep Neural Networks (DNNs) has pushed the state-of-the-art (SOTA) in speech recognition in a variety of settings \cite{pmlr-v48-amodei16,baevski2020wav2vec}. However, DNNs are resource-hungry, and building efficient ASR systems requires a lot of compute and supervision in the form of labeled data. Thus, recent Self-Supervised Learning (SSL) approaches \cite{baevski2020wav2vec,yang21c_interspeech,9585401,chen2021wavlm} which can learn representations from unlabeled audio data directly, have been gaining much traction. The primary aim of SSL is to use raw speech \cite{baevski2020wav2vec}, or other low-level features like Filter Banks \cite{liu2020mockingjay}, to learn high-level representations that prove to be effective in other downstream speech processing tasks. SSL has shown considerable performance boosts in building ASR systems, especially in settings where labeled data is scarce (as low as 10 minutes), and has been known to generalize better than supervised learning.

However,  several recent studies have highlighted the drawbacks of SSL. Firstly, the pretext tasks that the system is subjected to solve under the SSL paradigm are compute-intensive \cite{baevski2020wav2vec} and require a lot of unlabeled data. Secondly, a recent study reveals that similar to supervised learning, SSL too gets biased to the domain from which the unlabeled data originates \cite{hsu21_interspeech}. Thirdly,  as SSL implicitly learns a language model and other semantic information through the tasks it is subjected to solve \cite{pasad2021layer}, the generalizability of these models is only to the extent where data from a similar language or phonetic structure is introduced to it at the fine-tuning stage. Thus, as correctly pointed out by \cite{hannun2021history}, SSL for speech suffers from scale problems, and SSL generalizability can be improved with more efficient training procedures. Prior work for domain adaptation with self-supervised models mostly employ \emph{continued pre-training} or \emph{combined data pre-training} approaches \cite{hsu21_interspeech}. However, both assume the existence of high-resource unlabeled target-domain data, which is not always the case in a real-world scenario (for example, telephonic conversational speech is very difficult to procure due to privacy issues and not more than 1000 hours is freely available online).
%, and also involve SSL pre-training which is computationally intensive.

Building on the second and third problems mentioned above, in this paper, we try to address the problem of domain bias in pre-trained SSL models and try to devise an algorithm, that can allow pre-trained models trained on OOD data to easily adapt to the target domain and with improved performance, using only the supervised In-Domain data. To achieve this objective, we take the help of Unstructured Magnitude Pruning (UMP), wherein we select model parameters of ``least magnitude'', which we hypothesize to be of ``least importance'' to the domain, and propose to zero them out so that weights that are important for the downstream task should emerge with gradient updates, and those that are irrelevant should decrease in magnitude. This was first empirically proven in \cite{parp_mit}. The zeroed-out parameters are also kept trainable, which makes this different from generic DNN pruning, where the pruning mask does not allow gradient updates. More details about the different pruning strategies can be found in Alg.\ref{alg:cap}. We are motivated by the findings from the Lottery Ticket Hypothesis \cite{frankle2018the}, which suggests that a randomly initialized network contains sparse sub-networks which can be trained in isolation to match the result of a full model. Also, \cite{lu2022language} explores pruning for monolingual language adaptation in multilingual pre-trained models. To summarize, our main contributions are as follows:
\begin{itemize}
    \item We analyze the performance of different pruning strategies and frequencies for domain adaptation on pre-trained speech SSL models. We base our experiments on the practical assumption that only limited amounts of target-domain labeled data is available, and no other large-scale unlabeled corpus is available from the target domain.
    \item We propose \emph{Cross-Domain Task-Aware pruning} (CD-TAW), a first-of-its-kind method that uses readily available models fine-tuned on high-resource labeled OOD data to obtain the initial pruning mask relevant to the downstream ASR task. Experimental results reveal that CD-TAW performs better than generic fine-tuning, TAG, and TAW approaches.
\end{itemize}

\begin{figure} [t]
\centering
\includegraphics[width=\linewidth]{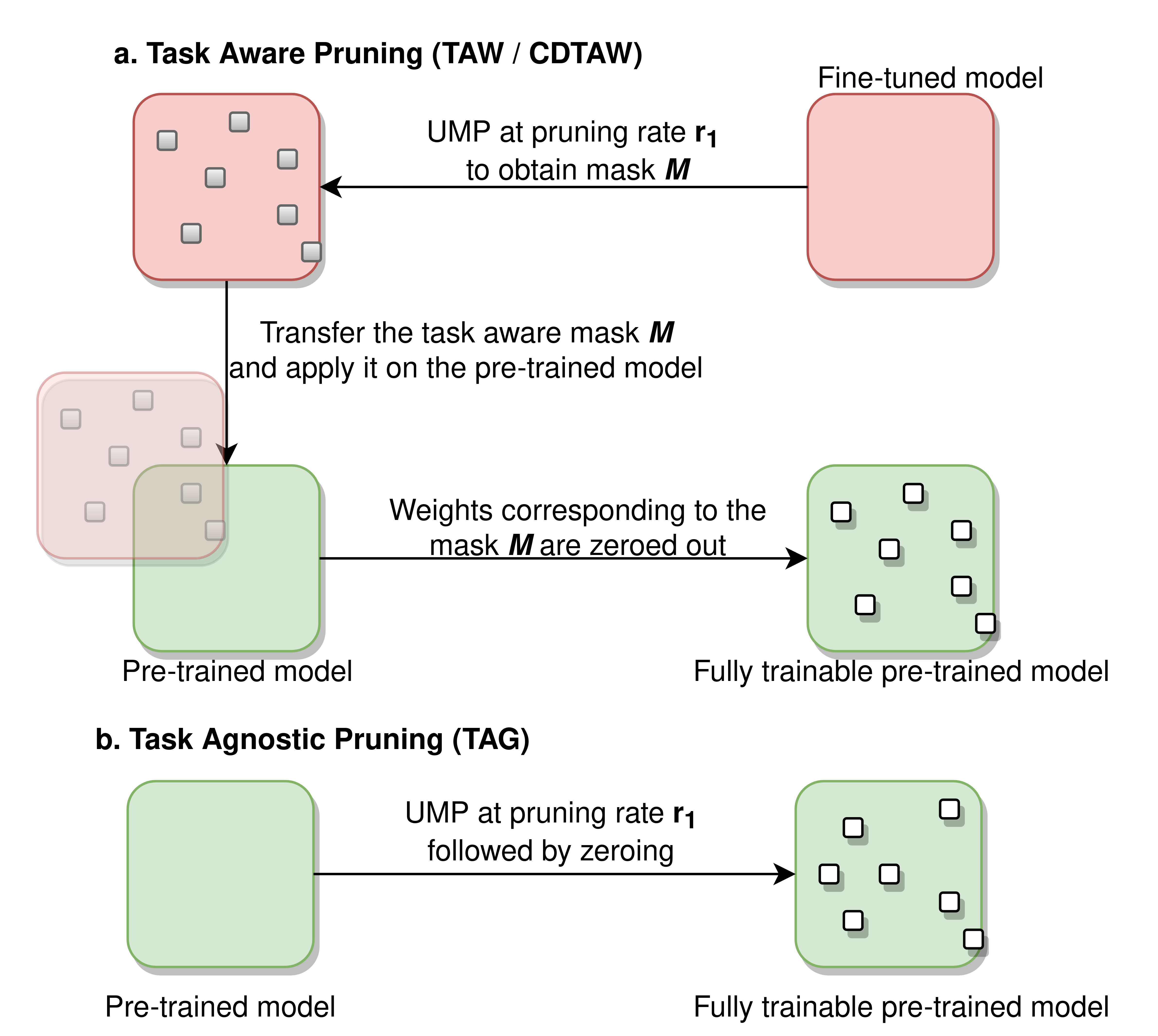}
\caption{The white blocks represent the weights that have been zeroed out.}
\label{fig:PADA_steps}
\end{figure}

\section{Related Work}

\subsection{Self-Supervised Speech Representation Learning}
SSL for speech representation learning has been explored primarily under 3 main paradigms, each solving a form of Masked Acoustic Modelling (MAM). The first and the most common in this space is based on Contrastive Predictive Coding (CPC), which minimizes the InfoNCE loss \cite{oord2018representation,baevski2020wav2vec}. The second paradigm learns by minimizing the Cross-Entropy loss, wherein the primary task is to predict the correct pseudo-label assigned to a frame, masked at the input to the model, leveraging the contextualized embedding of that frame obtained from a transformer encoder \cite{9585401,chen2021wavlm}. The third and the final paradigm solves the reconstruction-based objective function \cite{liu2020mockingjay,liu2021tera}.

\subsection{Pruning}
The concept of pruning DNNs has been extensively studied in the past \cite{frankle2018the,gale2019state,han2015learning}. Authors of \cite{h.2018to} show that large sparse models obtained through pruning generally perform better when compared to smaller but dense models. One of the first works that proposed pruning for domain adaptation was by \cite{gu-etal-2021-pruning}. However, they propose an entirely different methodology using knowledge distillation and employ it for text-based Neural Machine Translation (NMT). \cite{lu2022language} propose a pruning approach to improve monolingual ASR performance in multilingual SSL pre-trained models. Authors of PARP \cite{parp_mit} proposed the first work on pruning large self-supervised speech pre-trained models. They intend to find a sparse fine-tuned sub-network at some target sparsity level, with improved performance than using a full model. However, our work differs from theirs primarily from the perspective of the final task we try to solve. We approach iterative UMP \cite{frankle2018the} from a domain adaptation perspective and devise our algorithm by hypothesizing accordingly. Our end goal is to adapt the pre-trained SSL model to the target-domain, and the focus is not on obtaining a sparse model. One key difference from PARP is that our \emph{task-aware} pruning strategies work better for domain adaptation. At the same time, PARP resorts to \emph{task-agnostic} pruning strategies as they do not observe major differences between the two, given that they fine-tune on the data from the same domain as the pre-trained model's unlabeled data. Another key difference from PARP is that, while they progressively increase the pruning percentage to achieve the target sparsity in PARP-P, we dynamically decay the pruning percentage for better domain adaptation as explained in Section \ref{sec:algo}.
% For example, while PARP-Progressive proposed in \cite{parp_mit} iteratively increases target pruning sparsity with training steps, we decrease the same with training steps as a logical design choice for domain adaptation. Another 

\subsection{Domain Adaptation}

Domain Adaptation (DA) for building efficient ASR systems has been a well-studied topic in literature, with early work focusing on regularization \cite{yu2013kl,liao2013speaker}, teacher-student learning, \cite{meng2018adversarial,manohar2018teacher} or adversarial learning \cite{shinohara2016adversarial,meng2018speaker}. Lately, unsupervised domain adaptation of ASR models has been gaining traction, and researchers have been trying to find ways to use huge amounts of unlabeled data from the target domain for domain adaptation \cite{manohar2018teacher,anoop2021unsupervised,hwang2021large}. Continued pre-training is another common approach used \cite{gururangan-etal-2020-dont}. We emphasize that our work is one of the first to approach domain adaptation from a pruning perspective, involving no continued pre-training or OOD unlabeled data.

\section{Proposed Methodology}

\subsection{Problem Formulation}

Suppose we have a high-resource OOD unlabeled dataset $\textbf{P}$ and a medium-to-high resource OOD labeled dataset $\textbf{J}$, both from domain $D_1$. We also have a low-resource labeled dataset $\textbf{L}$. Let $p(\theta)$ be the neural network pre-trained using self-supervision on the dataset $\textbf{P}$ from domain $D_1$ and let $f(\theta_j)$ represent the resultant neural network after CTC based ASR fine-tuning has been performed on $p(\theta)$ on the dataset $\textbf{J}$ from domain $D_1$, such that $\theta$,$\theta_j \in \mathbb{R}^d$, represents the $d$ number of network parameters. As a part of this work, the models $p(\theta)$ and $f(\theta_j)$ that we use are made available in the public domain through the works on wav2vec-2.0 and fairseq \cite{baevski2020wav2vec, ott2019fairseq}.

Our primary aim is to adapt our model $p(\theta)$, to domain $D_2$ by just using the low-resource dataset $\textbf{L}$, with performance better than generic fine-tuning of $p(\theta)$ on $\textbf{L}$. We achieve this objective by identifying the weights having the least importance and zeroing them out. Weights relevant for the downstream ASR task on $\textbf{L}$ would emerge with gradient updates, and those irrelevant are zeroed out to facilitate better adaptation of the model. As we aim to get a model that delivers better performance without on any focus on achieving a target sparsity, weights that have been zeroed out remain trainable on $\textbf{L}$, thereby making use of the entire model capacity.

\subsection{Pruning Assisted Domain Adaptation (PADA)}

The generic pruning strategy in modern deep learning toolkits involves obtaining a mask \textbf{M} for a subset of model parameters, which intuitively results in multiplying by a value of zero in the forward pass step. Additionally, the mask does not allow gradient updates for these parameters during the backward pass. As part of this work, we explore different pruning strategies based on UMP. The focus of PADA is to discover the ``least important weights'' in a pre-trained SSL model $p(\theta)$. Unlike generic pruning, PADA does not retain the mask and lifts the mask after a round of UMP, followed by zeroing out the identified parameters, and keeps the parameters trainable. PADA can be achieved in multiple ways, primarily differing in the procedure of obtaining the initial pruning masks at the beginning of training and the rate of pruning employed at every training iteration. Fig. \ref{fig:PADA_steps} illustrates the three major pruning strategies. In the following four subsections, we briefly discuss the pruning strategies involved in PADA and the algorithm to implement the same.

% As mentioned earlier, we solve domain adaptation in SSL pre-trained models through a pruning assisted approach which we call PADA. There are different pruning strategies available as a part of PADA to discover the ``least important weights'' in a pre-trained SSL model $p(\theta)$. Unlike generic pruning, PADA keeps zeroes the weights and keeps them trainable which intuitively makes space for the new domain. PADA can be achieved in multiple ways, primarily differing in the procedure of obtaining the initial pruning masks at beginning of training and the rate of pruning employed at every training iteration. In the next 4 subsections, we discuss in brief the pruning strategies involved in PADA and the algorithm to implement the same.

\subsubsection{Task-Agnostic Pruning (TAG)}
\label{sec:tag}

In this approach we perform UMP at a pruning percentage $r_1$ directly on $p(\theta)$, and the $r_1$ percent of weights with the least magnitude are zeroed out, resulting in the model $p(\theta_0)$. This approach does not consider the downstream ASR fine-tuning that $p(\theta)$ would be subjected to. Weights that have been zeroed out are identified to be of least importance in terms of magnitude, only based on the pre-training task. One must note that pre-training is used to learn representations that serve multiple downstream tasks, and fine-tuning the model on specific downstream tasks and datasets could lead to a different set of weights gaining importance.

\subsubsection{Task-Aware Pruning (TAW)}\label{sec:taw}
The Task-Aware Pruning strategy aims to zero out those weights from $p(\theta)$, that are of least importance specific to the downstream ASR task performed on the dataset $\textbf{L}$. To this end, we first fine-tune $p(\theta)$ on $\textbf{L}$ which results in the fine-tuned model $p(\theta_f)$. We then perform UMP at a pruning percentage $r_1$ on $p(\theta_f)$, to obtain the mask $\textbf{M}$. The mask $\textbf{M}$ carries the information about the least important weights when $p(\theta)$ is exposed to the ASR downstream task on the dataset $\textbf{L}$. The exact same weights specified by the mask $\textbf{M}$ are zeroed out from $p(\theta)$, resulting in the model $p(\theta_0)$.

\subsubsection{Cross-Domain Task-Aware Pruning (CD-TAW)}\label{sec:cdtaw}
While TAW focuses on identifying and zeroing the weights of least importance specific to the downstream ASR task on the dataset $\textbf{L}$, it is evident that such an approach requires fine-tuning twice on the dataset $\textbf{L}$ (once to obtain the mask and the second time to fine-tune the pre-trained model carrying $r_1$ percent of zeroed weights). To alleviate this compute-heavy issue, we propose a method where UMP is performed at a pruning percentage $r_1$ on $f(\theta_j)$, to obtain the mask $\textbf{M}$. Similar to TAW, the exact same weights specified by the mask $\textbf{M}$ are zeroed out from $p(\theta)$, resulting in the model $p(\theta_0)$. The advantage of using $f(\theta_j)$ to obtain the mask $\textbf{M}$ is that, $\textbf{J}$ is a relatively high resource dataset (though OOD) in comparison with $\textbf{L}$. This results in the fine-tuning mask being more suited to the downstream ASR task, as $f(\theta_j)$ has seen more ``task'' related data. The reason behind mentioning that CD-TAW requires only one round of fine-tuning is that, when $f(\theta_j)$ is readily available online, it saves the explicit fine-tuning on the downstream ASR task to obtain a task-aware mask.

\subsubsection{Algorithm}\label{sec:algo}

Alg. \ref{alg:cap} shows the algorithm of PADA where we describe PADA with notations from earlier subsections.

We fix the initial pruning percentage $r_1$. We then choose the pruning frequency $P_{freq}$ from the 3 variants (Once / Iterative / Dynamic Iterative). We also fix the total number of fine-tuning updates to $N$.

\begin{algorithm}[!ht]
\caption{Pruning Assisted Domain Adaptation (PADA)}
\label{alg:cap}
\begin{algorithmic}[1]
\State Based on the pruning strategy chosen (TAG / TAW / CD-TAW), obtain the model $p(\theta_0)$ after zeroing out $r_1$ percentage of weights.
\If{$P_{freq}$ is Once}
\State Fine-tune $p(\theta_0)$ on $\textbf{L}$ for $N$ model updates to obtain the final domain adapted model $p(\theta_N)$.
\ElsIf{$P_{freq}$ is Iterative or Dynamic Iterative}
\State Choose the number of fine-tune updates $n$ after which you wish to re-prune and zero out some model weights.
\State Fix the pruning percentages $r_2, r_3, \cdots, r_k$.
\State Set $updates=0$ ; $n1=0$ and $u=1$.
\While{$updates \leq N$}
\State Train $p(\theta_{n1})$ on $\textbf{L}$, for $n$ updates to obtain $p(\theta_{n2})$.
\State Set $n1 = n2$ ; $updates = updates + n$.
\If{$updates \le N$ and $u < k$}
\State Perform UMP on $p(\theta_{n1})$ at a pruning percentage $r_u$ and zero out the pruned weights.
\State Set $u = u+1$.
\EndIf
\EndWhile
\EndIf
\end{algorithmic}
\end{algorithm}

When an Iterative PADA is used, frequency $r_i$ corresponds to $r_1 = r_1 = r_2 = \cdots = r_k$. However, when the Dynamic Iterative PADA is chosen, $r_1 > r_2 > r_2 > \cdots > r_k$. Dynamic Iterative PADA has the fundamental advantage in which fewer weights are zeroed out in the future training iterations where the model has ``already been trained on the target domain'' and needs ``lesser space'' to adapt. Precisely, each of the pruning strategies mentioned in sections \ref{sec:tag}, \ref{sec:taw}, \ref{sec:cdtaw} differ only in step 1 of the PADA algorithm. However, it is worth mentioning that step 1 of PADA is the most important step, which makes all the difference in terms of performance.

\section{Experimental Setup}

\subsection{Datasets and Pre-trained models}
The pre-trained models which we choose for PADA are:

\begin{itemize}
    \item \textbf{wav2vec-2.0 LV-60k}: A LARGE model \cite{baevski2020wav2vec} pre-trained on 60k hours of Libri-Light audio data.
    \item \textbf{wav2vec-2.0 LS-960}: A BASE model \cite{baevski2020wav2vec} pre-trained on 960 hours of LibriSpeech audio data.
    \item \textbf{XLSR-53}: A LARGE model pre-trained on 56k hours of data from 53 different languages \cite{babu2021xlsr}. The datasets used include Multilingual LibriSpeech (MLS), CommonVoice (CV), and Babel.
\end{itemize}
The target domain datasets we choose for PADA are:

\begin{itemize}
\item \textbf{Switchboard data} : It is a corpus of telephonic speech conversations \cite{godfrey1992switchboard}, a domain completely different from the Libri-Light data which is made up of read speech from audiobooks \cite{kahn2020libri}. We choose two subsets of the Switchboard data for our experiments, one of 30 hours and another of 2 hours, representing the low-resource and the extreme low-resource settings of domain adaptation. We report the word error rate (WER) results on the Switchboard dev set.

\item \textbf{Hindi Challenge Data} : It is a corpus of Hindi read speech data released as a part of the Hindi ASR Challenge\footnote{https://sites.google.com/view/asr-challenge}, which has 50 hours of speech data sourced from varied domains like politics, sports, entertainment etc. We choose a 7-hour subset of this data for the purpose of domain adaptation and report the character error rate (CER) results on the evaluation set released as a part of the challenge.
\end{itemize}

\begin{figure}[t]
  \centering
  \includegraphics[width=\linewidth]{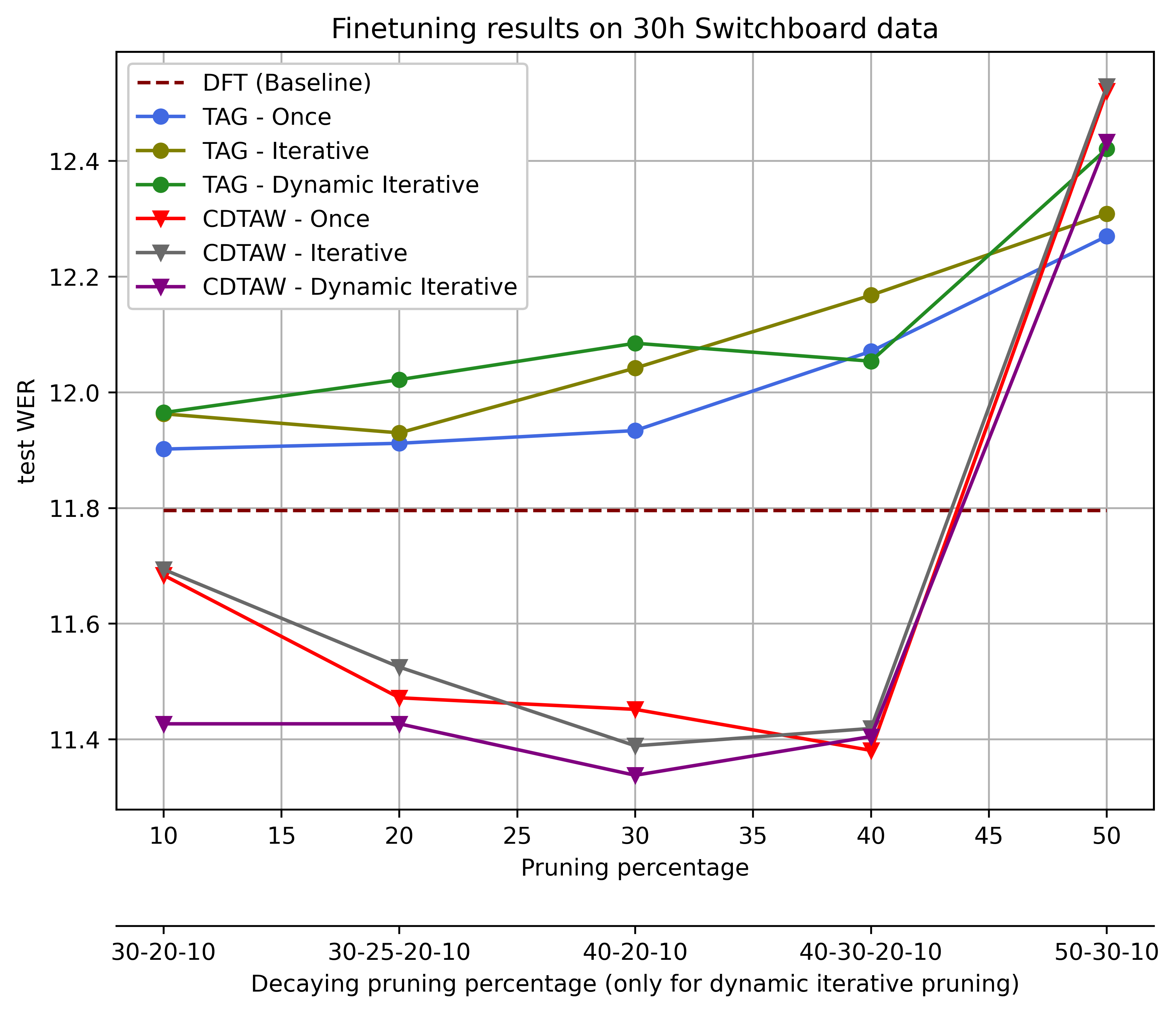}
  \caption{Comparison of various pruning strategies and frequencies on the wav2vec-2.0 large LV-60k model.}
  \label{fig:30h_results_plot}
\end{figure}

\subsection{Fine-tuning configuration}
The baseline approach chosen for comparison with PADA is the one where we directly fine-tune the pre-trained model on the target domain dataset $\textbf{L}$. This approach is referred to as Direct Fine-tuning (DFT), the baseline in Table \ref{tab:comp}.

We add a task-specific linear layer on top of the model and jointly fine-tune with CTC loss. As zeroing out weights is involved in PADA, we fine-tune the weights during each iteration on both DFT and PADA with no freezing of layers or updates being involved. The $N$ and $n$ values used in the Alg.\ref{alg:cap} are mentioned in the Table \ref{tab:comp}.

Extensive experimentation on the 30h Switchboard data, as depicted in Fig. \ref{fig:30h_results_plot}, helps us conclude with the best settings to go for, with the different pruning frequencies. The LARGE models use $r_1 = 40$, $r_1 = r_2 = r_3 = 30$, $r_1 = 40, r_2 = 20, r_3 = 10$ whereas the BASE models use $r_1 = 30$, $r_1 = r_2$ 

\renewcommand{\arraystretch}{1.1}
\begin{table*}[!ht]

  \caption{Performance of Direct Fine-tuning (DFT) compared to PADA with TAG, TAW, and CD-TAW.}
\label{tab:comp}
  \centering
  \resizebox{17.0cm}{!}{%
  \begin{tabular}{P{6cm} P{2.75cm} P{2.85cm} P{2.3cm} P{2.3cm} P{2.6cm}}
    \hline
    \textbf{Pre-trained model} & \textbf{DFT performance} & \textbf{Pruning Strategy} &
      \multicolumn{3}{c}{\textbf{Performance at different pruning frequencies}} \\
      \cline{4-6}
      &  &  & Once & Iterative & Dynamic Iterative\\
    \hline
    \textbf{Finetuning on 30h Switchboard (WER)}& & & & & \\
    
    \multirow{3}{4cm}{wav2vec-2.0 LV-60k\\ $N = 27000$\\ $n = 2400$}  &  & TAG   & 12.1 & 12.0 & 12.1 \\ \cline{3-6}
	& 11.8 & TAW  &  11.8 & 11.7 &  11.8 \\ \cline{3-6}
	&  & CD-TAW(LS-100h) & 11.4 & 11.4 & 11.3 \\ \cline{3-6}
	&  & CD-TAW(LS-960h)  &  \bf11.1 & \bf11.2 &  \bf11.0 \\ \hline \hline
	\textbf{Finetuning on 2h Switchboard (WER)}& & & & &\\
	
	\multirow{4}{4cm}{wav2vec-2.0 LV-60k\\ $N = 10000$\\ $n = 1000$}   &  & TAG   & 22.4 & 22.2 & 22.5 \\\cline{3-6}
	& 22.2 & TAW  &  21.2 & 21.3 &  21.3 \\ \cline{3-6}
	&  & CD-TAW(LS-100h) & 19.0 & 19.1 & 18.9 \\ \cline{3-6}
	&  & CD-TAW(LS-960h) & \bf17.6 & \bf18.2 & \bf17.6 \\ \hline \\
	
	\multirow{3}{4cm}{wav2vec-2.0 LS-960\\ $N = 10000$\\ $n = 1000$}   &  & TAG   & 29.8 & 29.9 & 29.8 \\\cline{3-6}
	& 29.2 & TAW  &  29.7 & 29.7 &  29.8 \\ \cline{3-6}
	&  & CD-TAW(LS-100h) & \bf27.4 & \bf27.3 & \bf27.3 \\\hline \hline
	
	\textbf{Finetuning on 7h Hindi data (CER)}& & & & & \\
	
	\multirow{3}{4cm}{wav2vec-2.0 LV-60k\\ $N = 9000$\\ $n = 1000$}   &  & TAG  & 18.7 & 14.1 & 13.6 \\\cline{3-6}
	& 16.3 & TAW  &  13.2 & 13.4 &  13.2 \\ \cline{3-6}
	&  & CD-TAW(LS-100h) & \bf13.2 & \bf13.4 & \bf13.0 \\\hline \\
	
	\multirow{3}{4cm}{XLSR-53\\ $N = 9000$\\ $n = 1000$}   &  & TAG & 12.0 & 11.9 & 12.0\\\cline{3-6}
	& 11.8 & TAW  &  \bf11.4 & \bf11.5 &  \bf11.3 \\ \cline{3-6}
	&  & CD-TAW(CV, Babel) & 12.0 & 11.9 & 12.0 \\
    \hline
\end{tabular}%
}

\end{table*}
\renewcommand{\arraystretch}{1.1}

$ = r_3 = 30$ and $r_1 = 30, r_2 = 25, r_3 = 20, r_4 = 10$ for the \emph{Once}, \emph{Iterative} and \emph{Dynamic Iterative} pruning frequencies respectively. The rest of the parameters follow standard configurations on fairseq \cite{ott2019fairseq} from the works of \cite{baevski2020wav2vec,babu2021xlsr} and are made available on GitHub\footnote{https://github.com/Speech-Lab-IITM/PADA} \footnote{Correspondence to Vasista Sai Lodagala: vasista.lodagala@gmail.com}.

% No LM was used while decoding to analyze the effect of PADA independent of the effect pf the LM.

\section{Results}

As evident from Table \ref{tab:comp}, CD-TAW implemented with wav2vec-2.0 LV-60k and the mask taken from 100hr LibriSpeech fine-tuning achieves a \textbf{20.6\%} relative WER improvement on the 2h Switchboard fine-tuning and \textbf{19.8\%} relative CER improvement on the 7h Hindi data fine-tuning.

\section{Result Analysis}

\subsection{Key observations}

In this section we try to elaborate and discuss on some of our key observations from our results in Table \ref{tab:comp}. They are as follows: 

\begin{itemize}
    \item \textbf{CD-TAW almost always outperforms TAW.} This corresponds to the fact that being more ``task-aware'' is more efficient while making parameters free to adapt to the new domain downstream than being ``domain-aware''. 
    \item \textbf{Dynamic Iterative Pruning} in most cases outperforms other fixed pruning frequency approaches. This re-confirms our hypothesis that, for domain adaptation, iteratively decreasing the pruning rate across fine-tuning iterations works well as the model gets more adapted to the new domain with time.
    \item \textbf{CD-TAW benefits with larger OOD labeled datasets.} We base this conclusion from the Table \ref{tab:comp} where CD-TAW on 2h Switchboard benefits more when the initial mask $\textbf{M}$ is taken from the model fine-tuned on 960h of LibriSpeech than 100h. It may seem unfair to compare masks coming from models fine-tuned on few hours of data with those fine-tuned on hundreds of hours of data. However, the point we are trying to emphasize is that, using such readily available models fine-tuned on large amounts of OOD data, brings better task-awareness and also avoids the need to fine-tune the models multiple times as in the case of TAW.
    \item \textbf{Beyond domain adaptation, PADA also helps in cross-language adaptation}. This is very evident from our experiments on the Hindi 7h labeled data. Using CD-TAW over TAG for fine-tuning on our low-resource Hindi dataset where the mask was taken from a LirbiSpeech fine-tuned model, gives us an absolute improvement of 5.5\% CER.
    \item \textbf{Language supervision impacts CD-TAW performance.} On fine-tuning the XLSR-53 model on the 7h Hindi data we notice that CD-TAW under-performs TAW as the amount of Hindi data in the OOD dataset (CV and Babel) is minimal. This is also in-lines with findings from \cite{lu2022language}.

\end{itemize}

% pruning frequency approaches for the reasons explained in \ref{sec:algo}.
\begin{figure} [t]
\centering
\includegraphics[width=\linewidth]{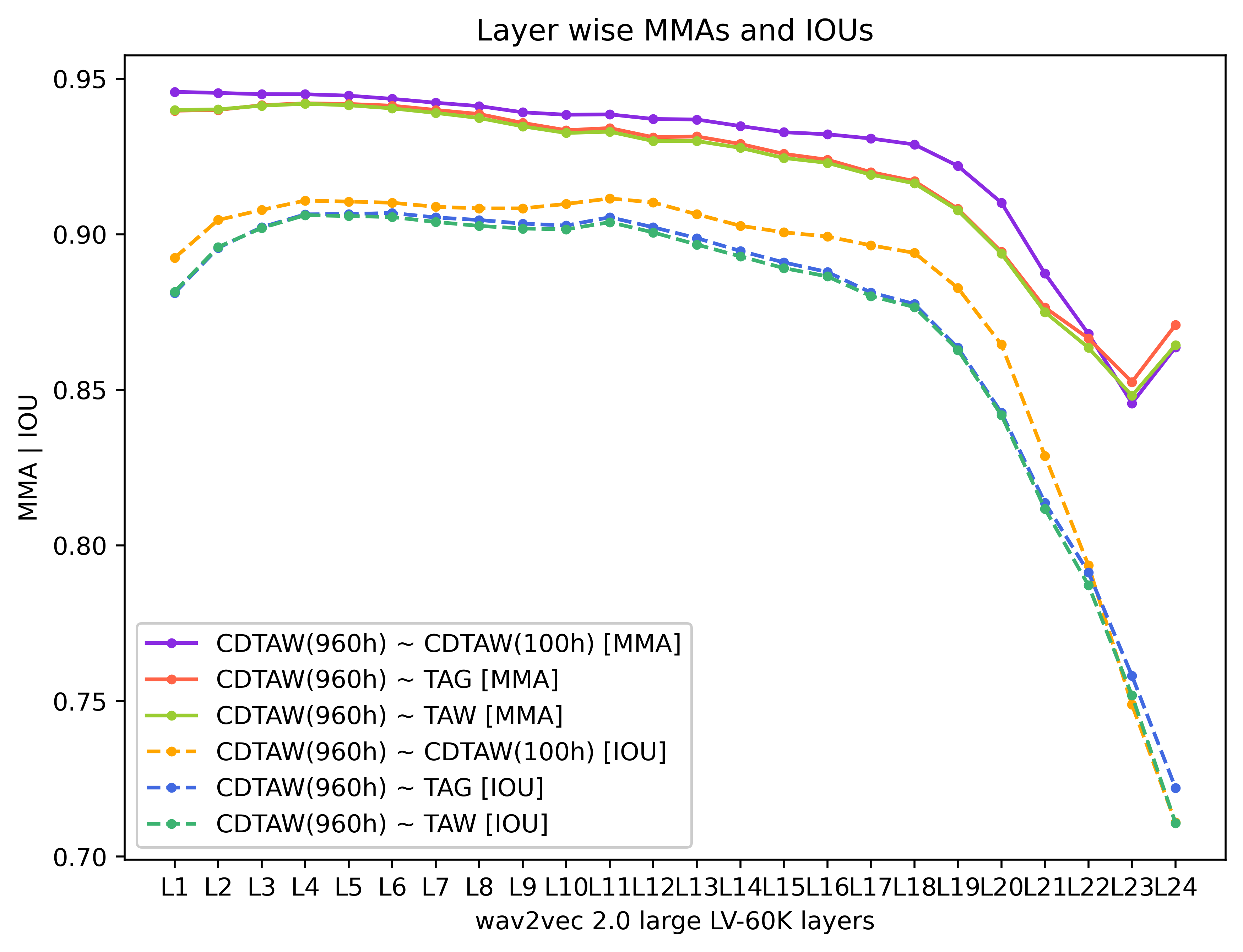}
\caption{Layerwise MMA and IOU for 2h Switchboard fine-tuning.}
\label{fig:Agreement and IOU}
\end{figure}

\subsection{Layer-wise Pruning Mask Comparison}

Next, we try to find the similarity between a mask that provides us with the best initial sub-network for fine-tuning our network on Switchboard using PADA, and all other masks from our other experiments. As evident from Table \ref{tab:comp}, our proposed CD-TAW provides the best initial sub-network mask when the mask is obtained from a converged model fine-tuned on 960hrs of LibriSpeech. To compare the similarity between masks, we use Intersection Over Union (IOU) and Mutual Mask Agreement (MMA) as the similarity measures. IOU for quantifying subnetwork masks, is defined in \cite{parp_mit} as follows:

\begin{equation}
    \operatorname{IOU}\left(m^{a}, m^{b}\right) = \frac{\left|\left(m^{a}=1\right) \cap\left(m^{b}=1\right)\right|}{\left|\left(m^{a}=1\right) \cup\left(m^{b}=1\right)\right|}
\end{equation}
where $m^{a}$ and $m^{b}$ are the two masks we want to compare.

We find that IOU scores are not completely reflective of the similarity between masks and define MMA as follows:
\begin{equation*}
    \operatorname{Agg_1}\left(m^{a}, m^{b}\right)= \left|\left(m^{a}=1\right) \cap\left(m^{b}=1\right)\right|
\end{equation*}
\begin{equation*}
    \operatorname{Agg_0}\left(m^{a}, m^{b}\right)= \left|\left(m^{a}=0\right) \cap\left(m^{b}=0\right)\right|
\end{equation*}

%  $\operatorname{Agg_1}\left(m^{a}, m^{b}\right)= \left|\left(m^{a}=1\right) \cap\left(m^{b}=1\right)\right|$.

% Also, let $\operatorname{Agg_0}\left(m^{a}, m^{b}\right) = \left|\left(m^{a}=0\right) \cap\left(m^{b}=0\right)\right|$.
\begin{equation}
    \operatorname{MMA}\left(m^{a}, m^{b}\right) = \frac{\operatorname{Agg_1} + \operatorname{Agg_0}}{N}
\end{equation}
where $N$ is the total number of weights in $m^a$ or $m^b$.

MMA provides a better comparison between masks compared to IOU because, while IOU focuses only in the unmasked regions, MMA measures the similarity between masks in both the masked and unmasked regions. For instance, let $m^a = [1, 0, 1, 0]$ and let $m^b = [1, 1, 0, 0]$. In this case though the two masks are similar in the first and last positions, the IOU score turns out to be $0.33$, while MMA gives the desired score of $0.5$.

Fig.\ref{fig:Agreement and IOU} depicts the Layer-wise MMA and IOU scores between our best sub-network mask and experiments from other masks. As clearly evident, both IOU and MMA fall sharply towards the last layers, which also learn the highest task-specific information \cite{pasad2021layer} (in our case, ASR task through CTC fine-tuning). Alternatively, our best-performing model, CD-TAW, re-confirms the hypothesis that irrespective of the domain, the more ``task-aware" the model is, the better the initial sub-network mask. Finally, though CD-TAW from a model fine-tuned on 100 hours performs worse than its 960hrs counterpart, it shows the highest IOU and MMA values out of all the masks, which again asserts that better masks can be obtained from more ``task-awareness''.

\section{Conclusion and Future Work}

In this paper, we propose PADA, a novel paradigm for low-resource ASR domain adaptation. Models pre-trained with SSL on high amounts of unlabeled OOD data show significant improvements when fine-tuned on ASR using the PADA framework. As part of the future work, we would like to explore if structured pruning and unsupervised SSL domain adaptation can help boost the performance in this learning paradigm.

\section{Acknowledgement}
A part of this work was funded by ``Bhashini: National Language translation Mission” project of the Ministry of Electronics and Information Technology (MeitY), Government of India.

\bibliographystyle{IEEEbib}
\bibliography{IEEE}

\end{document}